\documentclass[runningheads]{llncs}
\usepackage[T1]{fontenc}

\usepackage{cite}
\usepackage{amsmath,amssymb,amsfonts}
\usepackage{algorithmic}
\usepackage{graphicx}
\usepackage{textcomp}
\usepackage{xcolor}
\usepackage{bm}
\usepackage[ruled,vlined,linesnumbered]{algorithm2e}
\usepackage{amsmath}
\usepackage{enumitem}
\usepackage{booktabs}
\usepackage{multirow}
\usepackage{dcolumn}
\usepackage{graphicx,lipsum,afterpage,subcaption}
\usepackage{color}
\usepackage{amsmath}
\usepackage{hyperref}
\newcolumntype{C}[1]{>{\centering\let\newline\\\arraybackslash\hspace{0pt}}m{#1}}

\newcommand{\fact}[3]{$\textit{#2}(\textit{#1}, \textit{#3})$}
\newcommand{\mfact}[3]{\ensuremath{\textit{#2}(\textit{#1},\textit{#3})}}
\newcommand{\kb}{$\mathcal{K}$}
\newcommand{\mkb}{\ensuremath{\mathcal{K}}}
\newcommand{\V}{$\mathcal{V}$}
\newcommand{\mV}{\ensuremath{\mathcal{V}}}
\newcommand{\E}{$\mathcal{E}$}
\newcommand{\mE}{\ensuremath{\mathcal{E}}}
\newcommand{\I}{$\mathcal{I}$}

\newcommand{\PP}{$\mathcal{P}$}

\def\BibTeX{{\rm B\kern-.05em{\sc i\kern-.025em b}\kern-.08em
    T\kern-.1667em\lower.7ex\hbox{E}\kern-.125emX}}
\begin{document}

\title{Effects of Locality and Rule Language on Explanations for Knowledge Graph Embeddings\\
}

\author{Luis Galárraga\inst{1}\orcidID{0000-0002-0241-5379}}
\authorrunning{L. Galárraga}
\titlerunning{Effects of Locality and Rule Language on Explanations for KGE}
%
\institute{Inria\footnote{All affiliations: Inria, Univ. Rennes, CNRS, Irisa}, France \\
\email{luis.galarraga@inria.fr}}
\maketitle

\begin{abstract}
Knowledge graphs (KGs) are key tools in many AI-related tasks such as reasoning or question answering. 
This has, in turn, propelled research in
link prediction in KGs, the task of predicting missing
relationships from the available knowledge.  
Solutions based on KG embeddings have shown promising results in this matter. 
On the downside, these approaches are usually unable to explain their predictions. While some works have proposed to compute post-hoc rule explanations 
for embedding-based link predictors, these efforts have mostly resorted to rules with unbounded atoms, e.g., $\textit{bornIn}(x,y) \Rightarrow \textit{residence}(x,y)$, learned on a global scope, i.e., the entire KG. None of these works has considered the impact of rules with bounded atoms such as $\textit{nationality}(x,\textit{England}) \Rightarrow \textit{speaks}(x, \textit{English})$, or the impact of learning from regions of the KG, i.e., local scopes. We therefore study the effects of these factors on the quality of rule-based explanations for embedding-based link predictors. Our results suggest that more specific rules and local scopes can improve the accuracy of the explanations. Moreover, these rules can provide further insights about the inner-workings of KG embeddings for link prediction.
\end{abstract}

\section{Introduction}
\label{sec:introduction}
The continuous advances in information extraction on the Web have given rise to large repositories of machine-friendly statements modeled as knowledge graphs (KGs). These are collections of facts of the form \fact{s}{p}{o} that describe real-world entities, e.g., $\mfact{Italy}{capital}{Rome}$. In this formalism, the predicate $p$ in a statement \fact{s}{p}{o} can be seen as a directed labeled edge that connects the subject $s$ to the object $o$. KGs allow computers to ``understand'' the real world, and
find 
applications in multiple AI-related tasks such as entity-centric IR, reasoning, question answering, smart assistants, etc. Since KGs usually suffer from incompleteness, a central task in KGs is \emph{link prediction}, where the goal is to infer new facts from the available knowledge. Link prediction constitutes a fundamental step towards proper knowledge graph completion. 

Approaches for link prediction in KGs abound and fall mainly into two paradigms. On the one hand, \emph{symbolic methods}~\cite{pra,discovering_metapaths,hou2021rule} mine explicit patterns on the graph, e.g., the rule $\mfact{x}{capital}{y}$ $\Rightarrow \mfact{y}{inCountry}{x}$, and use those patterns to infer new relationships between entities. 
On the other hand, approaches based on latent factors~\cite{rescal,transe,hole,distmult,complex,ntn} embed predicates $p$ and entities $s,o$ in a latent space driven by a score function that ranks true facts better than false ones. For example, TransE~\cite{transe} learns $d$-dimensional embeddings (in bold) for predicates and entities such that $\bm{s} + \bm{p} \approx \bm{o}$, if \fact{s}{p}{o} holds in reality. TransE's score function for facts is then $-\lVert \bm{s} + \bm{p} - \bm{o}  \rVert_l$ ($l=\{1, 2\}$). 

Embedding-based methods have exhibited promising performance for link prediction, however their main downside is that they operate as black boxes: one cannot obtain an explanation of the logic behind a predicted fact \fact{s}{p}{o} from the latent representations of $s$, $p$, and $o$. This has therefore motivated some works on mining rule-based explanations for KG embeddings~\cite{excut,extracting-interpretable-models,towards-extracting-faithful-explanations,interpreting-embedding-models}. Those explanations can help us, for instance, verify if the embeddings meet expected reasoning guarantees such as  transitivity, i.e., 
$\mfact{x}{p}{z} \land \mfact{z}{p}{y} \Rightarrow \mfact{x}{p}{y}$, or detect biases in the data. It is known that redundancy in the form of inverse predicates, e.g., \fact{feline}{hyponym}{cat}, \fact{cat}{hypernym}{feline} in benchmark datasets, led to over-estimated accuracies for state-of-the-art embedding-based link predictors~\cite{meilicke2018,akrami20}. 
Had a mechanism to understand that the embeddings mainly captured patterns such as  
$\mfact{x}{hyponym}{y} \Rightarrow$ $\mfact{y}{hypernym}{x}$, this issue could have been detected in advance. 

A limitation of existing explanations for KG embeddings is that they only capture global inference patterns. This is tantamount to mining explanations in the language of unbounded atoms, i.e., rules with no constants in the arguments such as $\mfact{x}{bornIn}{z} \land \mfact{z}{officialLang}{y} \Rightarrow \mfact{x}{speaks}{y}$, that hold \emph{globally}, that is, on the entire KG. However, such rules cannot express specific entity associations such as $\mfact{x}{nationality}{USA} \Rightarrow \mfact{x}{speaks}{English}$, presumably captured by link predictors. 
On those grounds, Section~\ref{sec:evaluation} addresses the following research question (\textbf{RQ1}): \textbf{what is the impact of specific rules in the quality of the explanations for embedding-based predictors?}. Moreover, and in line with existing works in interpretable AI~\cite{lime,explanation-mining}, we also study a second research question (\textbf{RQ2}): \textbf{how does learning explanation rules on specific regions of the KG, i.e., local explanations, impact the quality of the resulting rules?}. Before answering these questions, we discuss basic concepts and related work in Section~\ref{sec:preliminaries}, and explain how to compute rule-based explanations for link predictors in Section~\ref{sec:approach}.

\section{Preliminaries}
\label{sec:preliminaries}
\subsection{Background Concepts}

\subsubsection{Knowledge Graphs. }
A knowledge graph \kb{} = (\V, \E, $l_v$, $l_e$) is a directed labeled graph with sets of vertices \V{} and edges \E{}, where the injective functions $l_v : \mV \rightarrow \mathcal{I}$,  $l_e : \mE \rightarrow \mathcal{P}$ assign labels to the vertices and edges. The sets \I{} and \PP{} contain entity and predicate labels. 
An edge labeled \emph{capital} departing from a vertex labeled \emph{France} to a vertex labeled \emph{Paris} denotes the \emph{statement} or \emph{fact} \fact{France}{capital}{Paris}, i.e., France has capital Paris. 
Hence, a KG $\mkb \subset \mathcal{I} \times \mathcal{P} \times  \mathcal{I}$ is also a set of facts \fact{s}{p}{o} with subject $s$, predicate $p$, and object $o$. Usually, standard KGs store only facts believed to be true. 

We define the \emph{potential set} $\Omega(\mkb)$ of a KG as the universe of facts that could be constructed from the entities and predicates in \kb. More formally, 
$\Omega(\mkb) =  \mathcal{D}_v(\mkb) \times \mathcal{D}_e(\mkb) \times \mathcal{D}_v(\mkb)$ where

\noindent
\begin{minipage}{0.49\linewidth}
	\[ \mathcal{D}_v(\mkb) = \{ l_v(v) : v \in \mV \}, \]
\end{minipage}
\begin{minipage}{0.49\linewidth}
 \[ \mathcal{D}_e(\mkb) = \{ l_e(e) : e \in \mE \} \] 
\end{minipage} \\ \\
 are the \emph{entity and predicate domains} of \kb. 
Furthermore, we define the \emph{potential set} of a predicate $p$ as 
$ \Omega(\mkb) \supseteq \Omega^p(\mkb) = \{ p(s, o) : (s, o) \in \mathcal{D}^p(\mkb) \times \bar{\mathcal{D}}^p(\mkb) \}$ 
with \\
\begin{minipage}{0.49\linewidth}
	\[ \mathcal{D}^p(\mkb) = \{s: \exists o : p(s, o) \in \mkb \}, \]
\end{minipage}
\begin{minipage}{0.49\linewidth}
	\[ \bar{\mathcal{D}}^p(\mkb) = \{o: \exists s : p(s, o) \in \mkb \}. \]
\end{minipage} \\

\noindent $\Omega^p(\mkb)$ therefore defines the set of all possible facts that could be constructed with the known subjects and objects of predicate $p$.


\subsubsection{Horn rules.} 
An \emph{atom} $A$ is a statement with constant predicate such that its subject and object arguments can be variables $v \in \mathbb{V}$ with $\mathbb{V} \cap \mathcal{I} = \emptyset$. If $A$ has only variable arguments, we say $A$ is \emph{unbounded}, otherwise it is \emph{bounded}.   
A \emph{Horn rule} $R$ is a statement of the form $\bm{B} \Rightarrow H$ where the \emph{body} $\bm{B}$ is a conjunction of atoms $\bigwedge_{1\le i \le n}{A_i}$, and $H$ is the head atom. For instance, the rule $\mfact{x}{parent}{z} \land \mfact{z}{nationality}{y} \Rightarrow \mfact{x}{nationality}{y}$ states that parents and children have the same nationality. These rules usually come with scores that quantify their precision. It is common to require atoms in rules to have at least one variable, be transitively connected, and form \emph{safe} rules, that is, ensure that the head variables occur also in the body. 
This condition guarantees that the head variables are universally quantified, allowing for concrete predictions via \emph{substitutions}. A substitution $\sigma : \mathbb{V} \rightarrow \mathcal{I}$ is a partial mapping from variables to constants, such that its application to atoms or rules replaces each variable with its corresponding constant in the mapping. For example, applying the substitution $\sigma = \{ x \rightarrow \mathit{\textit{Marie Curie}}, y \rightarrow \mathit{France}\}$ to the atom $A:\mathit{nationality}(x,y)$, gives a new atom $\sigma(A):\mathit{nationality}(\textit{Marie Curie},\mathit{France})$. We say a rule $R : \bm{B} \Rightarrow H$ \emph{predicts a fact} $A'$ in a KG \kb{}, denoted by $R \land \mkb \vdash A'$, iff 
$\exists\; \sigma : (\forall B \in \bm{B} : \sigma(B) \in \mkb) \land \sigma(H) = A'$. Put differently a rule predicts a fact $A'$ if there exist a substitution $\sigma$ that (i) maps each atom in the rule's body to a known KG fact, and (ii) maps the head atom to $A'$. If $R$ predicts a statement $A'$ and $A' \in \mkb$, we say that $R$ \emph{predicts $A'$ correctly}, i.e., the prediction is a known fact, and we use the notation $R \land \mkb \vDash A'$.

%
%

\subsubsection{Link Predictors.}

A \emph{link predictor} $f : \Omega(\mkb) \rightarrow \mathbb{R}$ is a function that scores the facts in the potential set of a KG, usually assigning higher values to true facts.  Link predictors are mostly used to answer queries of the forms \fact{s}{p}{?} or \fact{?}{p}{o}, in other words, queries that ask for the most likely subject or object of a statement given the other two components.
Embedding-based link predictors operate on latent representations for entities, predicates, and facts in $\Omega(\mkb)$. Hence, they actually have the form $f = \hat{f} \circ h$, where $\hat{f} : \mathbb{C}^k \rightarrow \mathbb{R}$\footnote{Most methods embed the entities in real spaces, i.e., in $\mathbb{R}^k$, but a few, e.g.,\cite{complex} resort to vectors of complex numbers. } is a function defined on a $k$-dimensional representation for facts, and $h : \Omega(\mkb) \rightarrow \mathbb{C}^k$ maps facts to $k$-dimensional vectors. If the semantics of the vector components are not understandable to humans, we say that $f$ is a black box. That is the case for pure embedding-based link predictors such as TransE~\cite{transe} or ComplEx~\cite{complex}.

\subsubsection{Explanations.} An explanation $E = \langle \mathcal{R}, g \rangle$ for a black-box link predictor $f : \Omega(\mkb) \rightarrow \mathbb{R}$ consists of a set $\mathcal{R}$ of Horn rules and a function $g: \mathcal{R} \rightarrow \mathbb{R}$ that attributes higher scores to rules that ``agree'' with $f$. A rule $R: \bm{B} \Rightarrow H$ agrees with $f$, if $R$ predicts a fact $A \in \Omega(\mkb)$ also predicted by $f$. This definition assumes the existence of a threshold $\theta$ such that $f(A) \ge \theta$ is interpreted as the black box also ``thinking'' that $A$ is true. Explanations can be of different scope, namely \emph{global} when they are learned on the potential set $\Omega^p(\mkb)$ of a predicate $p$, or \emph{local} when they are learned on smaller regions of $\Omega^p(\mkb)$ as explained in Section~\ref{sec:approach}.

\subsection{Related Work}
\label{sec:relatedwork}

\subsubsection{Link Prediction.}
This problem has received a lot of attention in the last 10 years with approaches lying on a spectrum from symbolic methods to embedding-based techniques. We refer the reader to~\cite{dd340ed9ec6646eba41af5e2f97f08bc} for a comprehensive survey. 
Symbolic techniques learn explicit patterns, e.g., arbitrary subgraphs, paths, association rules, Horn rules, etc., from KGs and use those patterns as features to predict missing links between entities~\cite{pra,discovering_metapaths,hou2021rule}. 
In contrast, the common principle of embedding-based methods is to model entities and predicates as elements in a latent space, where predicates characterize interactions between entity embeddings. Those interactions are modeled as geometrical operations, e.g., translation in TransE~\cite{transe} where $\bm{s} + \bm{p} \approx \bm{o}$ for true facts \fact{s}{p}{o} ($\bm{s}, \bm{p}, \bm{o} \in \mathbb{R}^d$), or rotation in RotatE~\cite{sun2018rotate}.
 More recent methods resort to neural architectures~\cite{convkb,10.1609/aaai.v33i01.33013060} that exploit the vicinity of entities in the graph to learn proper latent representations for both entities and predicates.   
  
In all cases, a scoring function -- implemented by minimizing a loss function -- guides the training of the embeddings, which are learned to yield high scores for true facts and low scores for false facts. The latter are obtained by corrupting the true facts in the KG -- a task of utter importance for the quality of the embeddings~\cite{10.1007/978-3-030-88361-4_24,zhang2020_negative_sampling}. 

Other methods combine the strengths of symbolic patterns and embeddings~\cite{boschin2022combining}. In~\cite{meilicke2021why}, the authors improve the accuracy of different state-of-the-art embedding-based link predictors by removing those predictions that are not backed up by any of the Horn rules learned on the data. This strategy is complemented with a combined ranking that takes into account the individual rankings given by the rules and the embeddings. Some approaches~\cite{uniker,itere,10.1007/978-3-030-88361-4_24} propose iterative algorithms that use rules and embeddings to produce better examples for subsequent training. In contrast, other methods~\cite{kg-embedding-preserving-softrules,end-to-end-diff-proving} instruct the embeddings to comply to explicit reasoning patterns, e.g., transitivity, $\mfact{x}{p}{z} \land \mfact{z}{p}{y} \Rightarrow \mfact{x}{p}{y}$.

\subsubsection{Explaining the Black Box. }

Unlike symbolic approaches, link predictors based on embeddings are black boxes. Hence, there have been some efforts to explain their logic by mining explicit patterns~\cite{extracting-interpretable-models,towards-extracting-faithful-explanations,interpreting-embedding-models} with attribution scores. Among these patterns, Horn rules are the most popular. The rules are extracted using state-of-the-art rule or path mining algorithms~\cite{amie3,pra,rudik-plus,scalekb}, whereas the attribution scores are learned via machine learning, e.g., linear or logistic regression in the spirit of standard explanation techniques such as LIME~\cite{lime}.
Nevertheless, none of these approaches exploits the power of Horn rules at its best. 
For instance,~\cite{extracting-interpretable-models,towards-extracting-faithful-explanations} mine rule explanations of up to two atoms, e.g., $\mfact{x}{bornIn}{y} \Rightarrow \mfact{x}{livesIn}{y}$, whereas DistMult~\cite{distmult} can only learn \emph{pure} paths such as $\mfact{x}{bornIn}{z} \land \mfact{z}{inCountry}{y} \Rightarrow \mfact{x}{nationality}{y}$.
Hence, none of these methods can induce explanations in the language of bounded atoms such as $\mfact{x}{nationality}{UK} \Rightarrow \mfact{x}{speaks}{English}$. 
Furthermore, all these endeavors mine global explanations. Embedding-based models can, though, be very complex and therefore hard to approximate in the general sense. 
Thus we explore the effects of bounded atoms and locality in the quality of the explanations. 


\section{Explaining KG Embeddings for Link Prediction}
\label{sec:approach}
Algorithm~\ref{alg:explainer} describes a generic procedure to compute rule-based explanations for a black-box link predictor $f$ trained on a KG $\mathcal{K}$, containing both true ($\mkb^+$) and corrupted facts ($\mkb^-$), in line with existing approaches~\cite{extracting-interpretable-models,towards-extracting-faithful-explanations,interpreting-embedding-models}. The rules are learned on a context $C$ consisting of facts of a given predicate $p$. We elaborate on the stages of the algorithm and the different ways to define the learning context.
\begin{algorithm}
		\caption{Build Explanation}
		\label{alg:explainer}
		\KwIn{link predictor $f : \Omega(\mathcal{K}) \rightarrow \mathbb{R}$ trained on $\mkb =\mkb^+ \cup \mkb^-$, context $C \subset \Omega^p(\mathcal{K})$}
		\KwOut{an explanation $E = \langle \mathcal{R}, g \rangle$ with set of rules $\mathcal{R}$, $g: \mathcal{R} \rightarrow \mathbb{R}$}
			
		$\hat{\mathcal{K}} := \emptyset$ \\
		\ForEach{$A := p(s, o) \in C$}{
			\If{$f(A) \ge \theta$}{
				$\hat{\mathcal{K}} := \hat{\mathcal{K}} \cup \{ p^f(s,o) \}$ \\
			}\Else{
				$\hat{\mathcal{K}} := \hat{\mathcal{K}} \cup \{ \neg p^f(s,o) \}$ \\
			}
		}
		$\mathcal{R} := \text{rule mining on}\; \mkb \cup \hat{\mkb} \; \text{for predicates} \; {p^{f},\neg p^{f}}$ \\
		\Return $\mathit{build\text{-}rule\text{-}based\text{-}surrogate(\mathcal{R}, \mkb, \hat{\mkb}, f)}$ \\

\end{algorithm}

\noindent \textbf{Binarizing the black box.} To learn Horn rules that mimic a black-box link predictor $f$, we need to convert $f$'s scores for facts into true or false verdicts. To this end, lines 2--6 label each fact in the context $C$ by computing $f$'s score and then applying a threshold to decide whether the fact is deemed true or not by $f$. This set $\hat{\mathcal{K}}$ of annotated facts is represented by the surrogate predicates $p^{f}$, $\neg p^f$. For instance the fact \fact{A. Einstein}{speaks$^f$}{English} means that $f$ ``thinks'' that Einstein speaks English. \\

\noindent \textbf{Rule Mining.} Line 7 in Alg.~\ref{alg:explainer} learns a set $\mathcal{R}$ of Horn rules of the forms $\bm{B} \Rightarrow p^{f}(s, o)$ and $\bm{B} \Rightarrow \neg p^f(s, o)$ with confidence scores from the original KG \kb{} and the black-box annotated context $\hat{\mathcal{K}}$. \\  

\noindent \textbf{Learning the explanation. }
Finally, line 8 uses the rules in $\mathcal{R}$ as features to learn a surrogate model $f_s : \mathbb{R}^{|\mathcal{R}|} \rightarrow \mathbb{R}$ that mimics the binarized $f$ and provides importance scores for the mined rules.
Given a statement $A = \bar{p}^{f}(s, o) \in \hat{\mkb}$ with $\bar{p}^f \in \{p^f, \neg p^f\}$, we encode $A$ as a vector $x_A \in \mathbb{R}^{|\mathcal{R}|}$ such that its $i$-th entry is set as follows:
\[ 
x_A[i] = \begin{cases} 
\mathit{sgn}(A)\times\mathit{conf}(R_i) & R_i \land (\mkb \cup \hat{\mkb}) \vDash A \\
-\mathit{sgn}(A)\times\mathit{conf}(R_i) & R_i \land (\mkb \cup \hat{\mkb}) \vdash A' \;\text{with}\; A' \neq A  \\
0 & \text{otherwise} 
\end{cases}
\]
\noindent Here $\mathit{sgn}(A)=1$ if $A = \bar{p}^{f}(s, o)$, otherwise $\mathit{sgn}(A)=-1$.  If a rule $R_i \in \mathcal{R}$ \emph{predicts correctly} a statement $A \in \hat{\mkb}$, the $i$-th component of $x_A$ holds a value equals the confidence of $R_i$ (reported by the rule mining phase) with the same polarity of $f$'s prediction. In that case, the rule $R_i$ agrees with $f$ and is a potential explanation for $f$'s answer on $A$. If $R_i$ is a potential explanation for some other fact $A'$, we change the sign of confidence value. In any other case, we assign a score of 0 to the entry.
We use the $x_A$ vectors and the binarized labels -- given by $\mathit{sgn}(A)$ -- to train a surrogate logistic regression classifier $f_s$, whose coefficients define an attribution mapping $g: \mathcal{R} \rightarrow \mathbb{R}$ for rules -- our explanation. The surrogate $f_s$ can provide both binary labels and probability scores for facts, and its coefficients can be used to rank the rules predicting true and false verdicts $p^f(s, o), \neg p^f(s, o)$. \\

\noindent \textbf{Explanation Context.} 
Existing explanation approaches for KG embeddings~\cite{interpreting-embedding-models,towards-extracting-faithful-explanations} 
mine global explanations, where
the context $C$ given as input to Alg.~\ref{alg:explainer} contains a large sample of true and false statements. The latter are obtained by corrupting the  true facts, so that for each fact $\mfact{s}{p}{o}$ we also add  $\{ p(s',o), p(s, o') \}$ ($s \neq s', o \neq o'$). The resulting surrogate $f_s$ approximates $f$'s general logic when predicting $p$-labeled links. 

A drawback of explanations based on global surrogates is that they assume that rules have always the same importance for all $p$-labeled predictions. Such a simplistic assumption can make explanation mining uninformative, if for example, 
the black box has a fine-grained behavior, i.e., it implements different logics for different regions of the KG. On those grounds, we propose to mine explanations within a \emph{local} scope obtained by calling Alg.~\ref{alg:explainer} on different sub-contexts $C' \subseteq C$ with triples that are close to each other in the latent space. These sub-contexts are obtained by applying agglomerative hierarchical clustering on $\bm{s} \oplus \bm{o}$, i.e., on the latent representation of pairs $s, o$ for true facts $\mfact{s}{p}{o} \in C$\footnote{$\oplus$ denotes concatenation; sub-contexts are corrupted to obtain counter-examples.}. We can also define \emph{per-instance} contexts around a target fact $A=p(s, o)$ by calling Alg.~\ref{alg:explainer} on a sub-context $C' = \{ A' = p(s', o) : A' \in C  \} \cup \{ A' = p(s, o') : A' \in C \} \cup \{ A \}$, that is, on statements that share at least one argument with $A$. 

\section{Evaluation}
\label{sec:evaluation}
To answer our research questions, we study the impact of bounded atoms (\textbf{RQ1}) and locality (\textbf{RQ2}) on the fidelity of rule explanations for embedding-based link predictors through a quantitative and an anecdotal evaluation.  

\subsection{Experimental Setup}
\label{subsec:experimental_setup}


\subsubsection{Datasets and Link Predictors.} We use the benchmark datasets \texttt{fb15k-237}, \texttt{wn18rr}, and \texttt{yago3-10}, on which we trained the bilinear methods ComplEx~\cite{complex} and HolE~\cite{hole}, and the translational approach TransE~\cite{transe}.
We used the implementations and data offered by the Torch-KGE library~\cite{arm2020torchkge}. 

\subsubsection{Rule Mining.} We mine Horn rules with AMIE~\cite{amie3}, a state-of-the-art rule miner for large KGs. 
By default, AMIE mines closed Horn rules\footnote{These are safe rules where each variable occurs in at least 2 atoms} of up to 3 unbounded atoms, but it can be instructed to mine longer rules, as well as to allow bounded atoms. Longer rules in combination with bounded atoms increase significantly the search space for rules, therefore we did not experiment with more than 3 atoms to avoid prohibitive runtimes~\cite{amie-plus}.
AMIE does not support explicit counter-examples to estimate the precision of rules, as required by Alg.~\ref{alg:explainer}, hence we extended the system to support explicit false facts in the precision computation. These counter-examples were generated through a variant of Bernoulli sampling that accounts for predicate domains~\cite{Wang_Zhang_Feng_Chen_2014}. We use all rules making at least 2 correct predictions with a precision of at least 10\% to learn the surrogate model (see Section~\ref{sec:approach}).   


\subsubsection{Explanations.} We compute rule-based explanations for the studied link predictors using the test instances of the experimental datasets to construct contexts $C$ of different scopes, i.e., global, local, and per-instance as explained in Section~\ref{sec:approach}. 
 For each call to Alg.~\ref{alg:explainer}, we split $C$ into
train and test sets $C_{\mathit{train}}$ and $C_{\mathit{test}}$ (30\%), so that we learn the explanations on $C_{\mathit{train}}$ and evaluate them on $C_{\mathit{test}}$. Local clusters are computed using agglomerative hierarchical clustering instructed to return $k$ clusters. We chose the most performing value of $k$ between 2 and 6.  

\noindent Link predictors are mainly used for two tasks: fact classification (true vs. false) and subject/object prediction for queries \fact{?}{p}{o} and \fact{s}{p}{?} where potential candidates are ranked by their score. We quantify the fidelity of our surrogate models (their ability to approximate the link predictors) for these two tasks via standard metrics, namely the ROC-AUC score and the mean reciprocal rank (MRR). The threshold $\theta$ to binarize $f$'s scores (line 6 in Alg.~\ref{alg:explainer}) is chosen via logistic regression as follows: we learned a logistic regression classifier $f_c: C \rightarrow [0, 1]$ using $f$'s scores on $C$ as input features and the real truth value of the facts as label -- corrupted facts are assumed false. In that spirit, $f_c$ estimates the probability that a given statement is true. $\theta$ is then chosen so that $f_c(\theta)=0.5$. 

\subsection{Results}
\label{subsec:quantitative_evaluation}


\begin{table*}[btp]
	\small
\centering
\addtolength{\leftskip} {-2cm} 
\addtolength{\rightskip}{-2cm}
\begin{tabular}{lC{0.6cm}C{0.6cm}C{0.6cm}C{0.67cm}C{0.67cm}C{0.67cm}C{0.6cm}C{0.6cm}C{0.6cm}C{0.67cm}C{0.67cm}C{0.67cm}C{0.6cm}C{0.67cm}C{0.6cm}C{0.6cm}C{0.67cm}C{0.6cm}}
	&\multicolumn{6}{c}{ROC-AUC}&\multicolumn{6}{c}{S-MRR}&\multicolumn{6}{c}{O-MRR} \\ \cmidrule(r){2-7} \cmidrule(r){8-13} \cmidrule(r){14-19} 
	&\multicolumn{3}{c}{Unbounded}&\multicolumn{3}{c}{Bounded}&\multicolumn{3}{c}{Unbounded}&\multicolumn{3}{c}{Bounded}&\multicolumn{3}{c}{Unbounded}&\multicolumn{3}{c}{Bounded} \\ \cmidrule(r){2-4} \cmidrule(r){5-7} \cmidrule(r){8-10} \cmidrule(r){11-13} \cmidrule(r){14-16} \cmidrule(r){17-19}    
	&\textbf{B}&L&PI&G&L&PI&\textbf{B}&L&PI&G&L&PI&\textbf{B}&L&PI&G&L&PI \\ \toprule
	complex&\underline{0.71}&0.68&0.64&0.93&0.93&\underline{\textbf{0.95}}&0.13&0.16&\underline{0.19}&0.31&0.35&\underline{\textbf{0.44}}&0.97&\underline{\textbf{1.00}}&0.97&0.97&\underline{0.98}&0.93\\ 
	transe&\underline{0.72}&0.70&0.64&\underline{\textbf{0.95}}&0.92&\underline{\textbf{0.95}}&0.12&\underline{0.20}&0.19&0.22&\underline{\textbf{0.47}}&0.45&0.97&\underline{\textbf{0.99}}&0.98&\underline{0.98}&0.93&0.91\\ 
	hole&\underline{0.66}&0.63&0.60&0.98&\underline{\textbf{0.99}}&\underline{\textbf{0.99}}&0.08&0.16&\underline{0.22}&0.27&0.36&\underline{\textbf{0.50}}&0.98&\underline{0.98}&0.97&0.98&\underline{\textbf{1.00}}&0.97\\ 
\end{tabular}
\caption{Fidelity on \texttt{fb15k-237}.  Best performances are in bold; best locality results are underlined. The baseline \textbf{B} are global explanations with unbounded atoms. G, PI, and L stand for global, per-instance, and local explanations. }
\label{tab:fb15k-237}
\end{table*}

\begin{table*}[btp]
	\centering
	\small
\addtolength{\leftskip} {-1.9cm} 
\addtolength{\rightskip}{-1.9cm}
\begin{subtable}{0.63\textwidth}

\begin{tabular}{lC{0.6cm}C{0.67cm}C{0.67cm}C{0.67cm}C{0.67cm}C{0.67cm}C{0.6cm}C{0.67cm}C{0.67cm}}
	&\multicolumn{3}{c}{ROC-AUC}&\multicolumn{3}{c}{S-MRR}&\multicolumn{3}{c}{O-MRR} \\ \cmidrule(r){2-4} \cmidrule(r){5-7} \cmidrule(r){8-10} 
	&G&L&PI&G&L&PI&G&L&PI \\ \toprule
	complex&0.55&0.64&\textbf{0.68}&\textbf{0.93}&0.60&0.38&0.92&0.93&\textbf{1.00}\\ 
	transe&0.51&0.55&\textbf{0.69}&0.71&\textbf{0.87}&0.32&0.93&0.92&\textbf{1.00}\\ 
	hole&--&0.65&\textbf{0.73}&--&\textbf{0.42}&0.38&--&\textbf{1.00}&\textbf{1.00}\\ 
\end{tabular}
\caption{\texttt{wn18RR}}
\label{tab:wn18RR}
\end{subtable}	
\hfill
\begin{subtable}{0.63\textwidth}
\begin{tabular}{C{0.67cm}C{0.67cm}C{0.6cm}C{0.67cm}C{0.6cm}C{0.6cm}C{0.6cm}C{0.67cm}C{0.67cm}C{0.67cm}}
	\multicolumn{3}{c}{ROC-AUC}&\multicolumn{3}{c}{S-MRR}&\multicolumn{3}{c}{O-MRR} \\ \cmidrule(r){1-3} \cmidrule(r){4-6} \cmidrule(r){7-9}  
	G&L&PI&G&L&PI&G&L&PI \\ \toprule
	0.55&\textbf{0.75}&0.00&\textbf{0.94}&0.39&0.17&0.87&1.00&1.00\\ 
	\textbf{0.66}&0.63&0.63&\textbf{0.73}&0.50&0.50&0.94&0.97&\textbf{1.00}\\ 
	0.71&\textbf{0.81}&0.65&\textbf{0.90}&0.38&0.26&0.93&\textbf{1.00}&0.99\\ 
\end{tabular}
\caption{\texttt{yago3-10}}
\label{tab:yago3-10}
\end{subtable}	
\caption{Fidelity of rule-based explanations with bounded atoms. }
\label{tab:wn18RRyago3}
\end{table*}


\subsubsection{Quantitative Evaluation.} Tables~\ref{tab:fb15k-237} and \ref{tab:wn18RRyago3} report the average ROC-AUC and MRR for the different explanation setups, namely unbounded vs. bounded rules learned on global (G), local (L), and per-instance (PI) scopes. The scores are computed by averaging the fidelity obtained for each call to Alg.~\ref{alg:explainer} weighted by the size of the corresponding test set, i.e., $|C_{\mathit{test}}|$. We disaggregate the MRR into S-MRR and O-MRR -- when the task is to predict the subject or object given the other two components. 

Our baseline setting (denoted by \textbf{B}) are global unbounded rules as mined by existing approaches~\cite{extracting-interpretable-models,towards-extracting-faithful-explanations,interpreting-embedding-models}. 
We highlight that we could not mine explanations with such a setting for \texttt{wn18RR} and \texttt{yago3-10} on any of the studied link predictors -- not even for local or per-instance scopes. This happens because unbounded rules can only be extracted when the training KG contains very prevalent and general regularities in the interactions between the predicates. 
The datasets \texttt{wn18RR} and \texttt{yago3-10}, however, have much fewer predicates than \texttt{fb15k-237}: 11 and 37 for the former versus 237 for the latter. 
Bounded atoms also increase the coverage explanation for \texttt{fb15k-237}. While the baseline provides explanations for 18 different predicates for ComplEx on \texttt{fb15k-237}, allowing bounded rules increases the coverage to 58 predicates (HolE and TransE exhibit comparable increases).
Moreover the results in Table~\ref{tab:fb15k-237} suggest that bounded atoms in rules generally increase fidelity. 

It is important to remark that allowing constants in the rule atoms comes at the expense of longer runtimes and many more, potentially noisy, rules. On \texttt{fb15k-237} with global scopes, for example, the number of unique rules mined from TransE increases from 1k to 193k. That said, only 134k of those rules get non-zero coefficients during the attribution phase -- implemented via logistic regression. This phase is indeed designed to identify the rules with actual explanation power w.r.t the veredicts of the link predictor. 

We also observe that rule-based surrogates tend to be better at mimicking the link predictors for object prediction. This is explained by the nature of KG predicates, which are usually defined in a subject-oriented manner, e.g., \fact{J. Biden}{nationality}{USA} and not \textit{hasCitizen}(\textit{USA}, \textit{J. Biden}). This makes subject prediction generally harder to mimic, because, e.g., it is easier to predict the nationalities of J. Biden than to predict all USA citizens. Besides, this phenomenon is  corroborated by the actual performances of the link predictors. For instance, ComplEx exhibits an average S-MRR of 0.29 on \texttt{wn18RR}, whereas the average O-MRR reaches 0.46. 
That said, Table~\ref{tab:wn18RR} suggests that S-MRR fidelity can still be high even in the presence of subject-oriented predicates.

When we look at the effects of locality on fidelity, we notice mixed effects. On \texttt{fb15k-237}, locality hurts ROC-AUC performance for unbounded rules and brings moderate performance gains for the MRR. This is probably because good general unbounded rules require more ``diverse'' data. The situation is different for bounded rules, for which locality boosts fidelity in most cases. These results suggest that locality and bounded rules are complementary.  
A similar behavior can also be observed for coverage. For example, local scopes combined with bounded rules allow mining 507k unique rules (with non-zero attribution) for 130 different predicates for ComplEx on \texttt{fb15k-237} vs. 112k rules/58 predicates and 1505 rules/62 predicates when only one of the features is enabled (the baseline mines 730 predicates covering 18 predicates). 
For per-instance scopes we can compute rule explanations for up to 2782 individual facts (out of 20k) covering 83 predicates (HolE on \texttt{fb15k-237}).
\begin{table}[hbtp]
	\centering
	\begin{tabular}{cc} \\ \toprule
		\multicolumn{2}{c}{\texttt{fb15k-237}} \\ \toprule
		\multicolumn{2}{c}{(1) $\mfact{x}{place\_of\_birth}{Chicago} \Rightarrow \mfact{x}{nationality}{USA}$ \textbf{[TransE]}}\\ 
		\multicolumn{2}{c}{(2) $\mfact{x}{has\_lived\_in}{Brooklyn} \Rightarrow \mfact{x}{nationality}{USA}$ \textbf{[ComplEx]}}\\
		\multicolumn{2}{c}{(3) 
			$\mfact{x}{profession}{Author} \Rightarrow  \mfact{x}{gender}{M}$ \textbf{[ComplEx]}} \\
		\multicolumn{2}{c}{(4) $\mfact{z}{impersonates}{x} \land \mfact{z}{gender}{y} \Rightarrow \mfact{x}{gender}{y}$ \textbf{[ComplEx, HolE]}}\\
		\multicolumn{2}{c}{(5) $\mfact{z}{country}{y} \land \mfact{x}{birth\_place}{z} \Rightarrow \mfact{x}{nationality}{y}$}\\ 
		\multicolumn{2}{c}{(6) $\mfact{z}{company}{x} \land \mfact{z}{athlete:sport}{y} \Rightarrow \mfact{x}{sport}{y}^\circ$ \textbf{[HolE]}} \\
		\multicolumn{2}{c}{(7) $\mfact{z}{fwc:club}{x} \land \mfact{z}{sport}{y} \Rightarrow \mfact{x}{sport}{y}^\circ$ \textbf{[HolE]}} \\
		\midrule
		\multicolumn{2}{c}{\texttt{yago3-10}} \\ \toprule 
		\multicolumn{2}{c}{(8) $\mfact{x}{affiliation}{Umeå IK} \Rightarrow \mfact{x}{gender}{F}\dagger$ \textbf{[TransE, ComplEx]}}\\
		\multicolumn{2}{c}{(9) $\mfact{x}{wonPrize}{O. Orange-Nassau} \Rightarrow \mfact{x}{wonPrize}{D.S. Medal}$ \textbf{[ComplEx]}}\\ \midrule
		\multicolumn{2}{c}{\texttt{wn18rr}} \\ \toprule 
		\multicolumn{2}{c}{(10) $\mfact{Insecta}{meronym\_mb}{x}  \Rightarrow \mfact{x}{hypernym}{Animal}\dagger$ \textbf{[ComplEx, HolE]}} \\ \bottomrule
	\end{tabular}
	\caption{Some rule explanations. $^\circ$, $\dagger$ denote local and per-instance explanations.}	
	\label{tab:examples}					   
\end{table}

\subsubsection{Anecdotal Evaluation.} Table~\ref{tab:examples} shows a few examples of rule-based explanations for our experimental link predictors. These correspond to some of the best ranked rules according to the coefficients of the surrogate classifiers. The rules illustrate regularities preserved by the link predictors, since the body of the rules defines conditions satisfied by the facts of the KG, in contrast to the head that matches statements predicted by our black boxes (see Alg~\ref{alg:explainer}). Rules with bounded atoms offer legible insights about the information that the link predictors may be capturing to make predictions. 

A key observation is that the different link predictors do not seem to rely on the same information -- as suggested by rules (1) and (2) for ComplEx and TransE on \texttt{fb15k-237}. This is supported by the fact that among the 47 predicates for which ComplEx finds global explanations with bounded atoms, only 3 have common rules with TransE. 
We bring our attention to rule (3), which suggests that embeddings do reproduce the biases in the source data\footnote{Rule (8),  on the other hand, refers to a women's football team.}. Recall that \texttt{fb15k-237} was mainly extracted from Wikipedia, known to have gender biases~\cite{wagner2016women}. Those biases are easier to spot with rules with bounded atoms, which are a complement to more general explanations such as (4) and (5).   
 We also highlight that local contexts can illustrate the semantics captured by the embeddings. This is exemplified by rules (6) and (7) that were learned on the same predicate but on two fact clusters. As we can see, our mining routine learned semantically equivalent rules, defined on different thematic domains, namely \emph{athletes} and \emph{fwc}; the latter refers to the 2010 FIFA World Cup.

\section{Conclusion}
\label{sec:conclusion}
We have studied the effects of specific rules with bounded atoms and local scopes on the quality of explanations for embedding-based link predictors on knowledge graphs. Our results suggest a rather positive impact on the explanation fidelity and the coverage of the explanations. Moreover, specific rules and local scopes exhibit a symbiotic relationship. 

Even though rule-based explanations reflect regularities preserved by black-box link predictors, they do not shed light on causality. In this line of though, we envision to compute causal explanations that help us understand the role of the different entities, predicates, and latent components of KG embeddings in the resulting predictions. We have also planned to elaborate more on the relationship between link prediction performance and explanation fidelity, in particular at the level of the individual predicates. 
The source code and experimental data of our work is available at \url{https://gitlab.inria.fr/glatour/geebis}. \\

\noindent \textbf{Acknowledgment.} This research was supported by TAILOR, a project funded by EU Horizon 2020 research and innovation programme under GA No. 952215.

\bibliographystyle{plain}
\bibliography{references}

\end{document}